 \documentclass[1+,12pt]{elsarticle}

\usepackage{amssymb}
\usepackage{amsthm}
\usepackage{enumitem}
\usepackage{booktabs}
\usepackage{bm}
\usepackage{multirow}
\usepackage{enumitem}
\usepackage{bbding}
\usepackage{makecell}
\usepackage{subcaption}
\usepackage{float}
\usepackage{amsmath}

\begin{document}

\begin{frontmatter}



\title{Self-Paced Neutral Expression-Disentangled Learning for
Facial Expression Recognition\tnoteref{t1}}
\tnotetext[t1]{This work was supported in part by Sichuan Science and Technology Program (Nos. 2021YFS0172, 2022YFS0047, and 2022YFS0055), Medico-Engineering Cooperation Funds from University of Electronic Science and Technology of China (No. ZYGX2021YGLH022), and Opening Funds from Radiation Oncology Key Laboratory of Sichuan Province (No. 2021ROKF02). Lifang He was supported by Lehigh’s accelerator grant S00010293.}

\author[1]{Zhenqian Wu}
\ead{russius@163.com}
\author[1]{Xiaoyuan Li}
\ead{lilan1196769290@163.com}
\author[1]{Yazhou Ren\corref{cor1}}
\ead{yazhou.ren@uestc.edu.cn}
\author[1]{Xiaorong Pu}
\ead{puxiaor@uestc.edu.cn}
\author[1]{Xiaofeng Zhu}
\ead{seanzhuxf@gmail.com}
\author[2]{Lifang He}
\ead{lih319@lehigh.edu}

\cortext[cor1]{Corresponding author}

\affiliation[1]{organization={School of Computer Science and Engineering, University of Electronic Science and Technology of China},
             city={Chengdu 611731},
             %%postcode={611731},
             country={China}}
             
\affiliation[2]{organization={Department of Computer Science and Engineering, Lehigh University},
             city={Bethlehem},
             %%postcode={18015},
             state={PA 18015},
             country={USA}}


%%%%%%%%% ABSTRACT
\begin{abstract}
The accuracy of facial expression recognition is typically affected by the following factors: high similarities across different expressions, disturbing factors, and micro-facial movement of rapid and subtle changes. One potentially viable solution for addressing these barriers is to exploit the neutral information concealed in neutral expression images. To this end, in this paper we propose a \underline{S}elf-\underline{P}aced \underline{N}eutral Expression-\underline{D}isentangled \underline{L}earning (SPNDL) model. SPNDL disentangles neutral information from facial expressions, making it easier to extract key and deviation features. 
Specifically, it allows to capture discriminative information among similar expressions and perceive micro-facial movements. In order to better learn these neutral expression-disentangled features (NDFs) and to alleviate the non-convex optimization problem, a self-paced learning (SPL) strategy based on NDFs is proposed in the training stage. SPL learns samples from easy to complex by increasing the number of
samples selected into the training process, which enables to effectively suppress the negative impacts introduced by low-quality samples and inconsistently distributed NDFs. Experiments on three popular databases (\emph{i.e.}, CK+, Oulu-CASIA, and RAF-DB) show the effectiveness of our proposed method.

\end{abstract}


%\item Neutral information in neutral expression is utilized for %facial expression recognition.
%\item Deviation information is extracted from facial movements while %disturbance information is removed.
%\item A self-paced learning strategy based on neutral %expression-disentangled features is proposed.
%\item A variation of our proposed method is adopted for in-the-wild %databases. 


\end{frontmatter}


\section{Introduction}
Facial expression is an important indicator of our psychological and physical state, making facial expression recognition (FER) a subject of concern. Due to high similarities across different expressions \cite{2021Feature}, disturbing factors \cite{2020Ruan}, and micro-facial movements that are difficult to perceive \cite{WEI2021159}, accurate and reliable results of FER are difficult to achieve.
\par
%The previous methods have made significant efforts on FER, most of them pay attention to one of the two orientations: disturbance-disentangling and effective features extracting. 

In recent years, deep learning has been widely applied in FER and has demonstrated impressive performance, which mainly focus on two orientations: disturbance-disentangling and effective feature extracting. 
The former one aims to eliminate the impact of disturbing factors such as illumination and identity \cite{2017Identity,2020Ruan,RAN}. The latter one focuses on the process of extracting expression features, trying to capture fine-grained features \cite{2021Feature}, transformed deep and shallow features \cite{BOUGOURZI2020113459}, expressive information \cite{2018Yang} or dynamic information \cite{2012Atlases,2014Liu,KHAN2017427}.      

\par
Despite the huge success and promise, existing FER methods generally ignore the fact that neutral information hidden in neutral facial expression is shared by both facial expression image and corresponding neutral expression image.
Here, neutral information includes initial state of expression and disturbance information from disturbing factors. Disturbance information can confuse networks during training stage, resulting in that the networks misjudge targets as other similar expressions, \emph{e.g.}, appearance characteristics such as wrinkles are usually mistaken for facial movements. Their redundancy will also reduce the weight of micro expression features, making micro-facial movements more difficult to perceive, \emph{e.g.}, the subtle motions of micro-facial expression are erased during the convolution process, so they play minor parts in the final features extracted by networks. 

\begin{figure}
\centering
\includegraphics[scale=0.6]{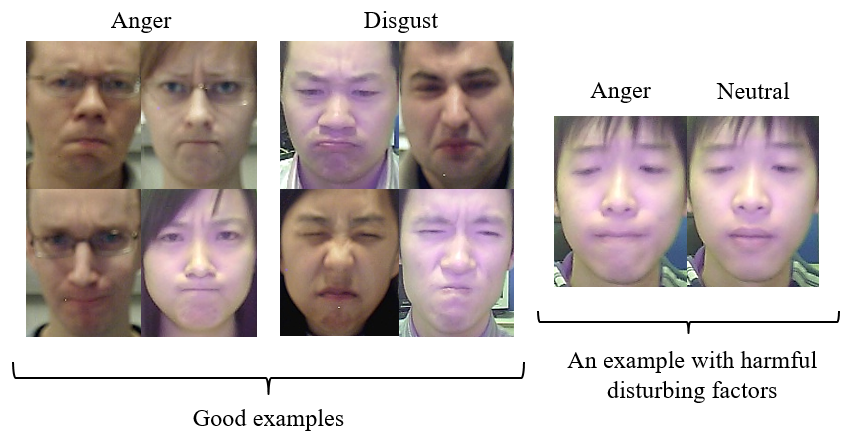}
\caption{Examples to show the harm caused by disturbance information. Images are from the Oulu-CASIA dataset.}
\label{fig:Examples}
\end{figure}

\par

Figure \ref{fig:Examples} shows an example with harmful disturbance information. In the left side of good examples, we can see that bulging mouth is an important characteristic of angry expressions while squinting eyes and pouting are distinguished characteristics of disgust expressions. 
%Then, given the example with harmful disturbing factors, 
In the right side of the example with harmful disturbing factors, the young man's anger face is easily misidentified as “disgust” because of his “squinting eyes”. 
%So the anger face on the right can easily be misidentified as “disgust” because of its “closed eyes”. 
However, according to his neutral face, “squinting eyes” is not his facial movement but his identity information of small eyes. It can be seen that facial expression image (\emph{e.g.}, anger face) and neutral expression image from the same person share similar disturbance information.

\par 
In contrast to disturbance information, initial state of expression only exists in neutral expression, but plays a favorable role in FER. With its help, we can extract deviation information and capture features from facial movements (such as raising eyebrows and slightly widening eyes) that may be easily covered up by identity. 
\par
Inspired by the above observations, in this paper we propose a \underline{S}elf-\underline{P}aced \underline{N}eutral Expression-\underline{D}isentangled \underline{L}earning (SPNDL) model, which exploits neutral facial expression to remove disturbance information and obtain initial state of expression.
%we aim at exploiting neutral facial expression to remove disturbance information and obtain initial state of expression. To this end, in this paper we propose a \underline{S}elf-\underline{P}aced \underline{N}eutral Expression-\underline{D}isentangled \underline{L}earning (SPNDL) model. 
We first employ a backbone convolutional neutral network (CNN) to extract basic features of input paired images, consisting of the facial expression image (called target image) to be predicted and its corresponding neutral expression image (called reference image).
%Besides the facial expression image to be predicted, called target image by us, its corresponding neutral expression image is chosen as reference image. 
These paired images are input to the same backbone network simultaneously, outputting two paired basic features.
Because the target image and its reference image share similar disturbance information, a simple but effective subtraction operation is applied to these paired basic features to obtain neutral expression-disentangled features (NDFs).
As initial state of expression exists in reference image, and peak state of expression exists in target image, the obtained NDFs contain the deviation information. 
In particular, to alleviate the distribution difference between two paired basic features, we apply a normalization operation between the output features of target image and reference image in each layer of backbone CNN. 
\par
Deep learning based FER methods solve non-convex optimization problems and are easily stuck into poor local solutions.
To alleviate this issue, we propose a SPL strategy for FER, which learns samples with easy NDFs first, and then gradually adapts to hard ones. This strategy can reduce the impacts of low-quality samples and paired samples with inconsistently distributed NDFs.
The main contributions of this paper can be summarized as follows:
\begin{enumerate}[label=\arabic*), leftmargin=*]
    \item We propose a neutral expression-disentangled method to remove disturbance information from facial expressions and capture the deviation information of facial movement, which is low-cost and can effectively improve the performance of facial expression recognition.
     \item We propose a self-paced learning strategy to alleviate the non-convex optimization issue, and simultaneously suppress the negative impacts of inconsistently distributed NDFs and low-quality data.
     \item Our method has no hyperparameters beyond those of deep learning models, making it easy to be trained. We achieved 99.69$\%$, 90.14$\%$ and 89.08$\%$ recognition accuracies respectively on three popular databases (\emph{i.e.}, CK+, Oulu-CASIA, and RAF-DB), which outperforms the state-of-the-art FER methods on the first two databases and gains competitive performance on the last database.  
\end{enumerate}

\section{Related Works}
This section reviews related methods for facial expression recognition and self-paced learning.
\vspace{0.2cm}
\par
\textbf{Facial Expression Recognition.} Over the past decades, classical facial expression classification techniques such as neural networks \cite{1998Zhang, 2004Tian}, support vector machine \cite{2005Bartlett}, bayesian network \cite{2002Cohen} and rule-based classifiers \cite{2000Pantic, 2004Pantic, 2006Dynamics} have been well proposed. Especially as the rise of the CNN architectures \cite{krizhevsky2012imagenet, szegedy2015going}, excellent performance has been shown on facial related recognition tasks \cite{yim2015rotating, zhu2013deep}. With the development of deep learning, researchers are devoted to obtaining effective expression information. Ruan \emph{et al}. \cite{2021Feature} view the expression information as the combination of the shared information across different expressions as well as unique information. Ding \emph{et al}. \cite{2016FaceNet2ExpNet} propose a novel FaceNet2ExpNet for the training of expression recognition network based on static image, where a two-stage training algorithm is designed by first using face net for convolution layers recognition in expression net and then refining it with fully connected layers. In dynamic expression recognition, Liu \emph{et al}. \cite{2014Liu} solve the temporal alignment and semantics-aware dynamic representation issues by manifold modeling of videos based on a novel mid-level representation. Zhao \emph{et al}. \cite{2016Peak} enable the natural evolution of expressions from non-peak to peak during the learning process using a novel peak-piloted deep network architecture. Yang \emph{et al}. \cite{2018Yang} generate subject's neutral face using generative adversarial networks, and then extract expressive features embedded in the encoder and decoder. 

In terms of disturbance disentangling, it is widely used to eliminate disturbing factors such as identity and pose in order to get the holistic expression features of images. Meng \emph{et al}. \cite{2017Identity} ameliorate the facial variety variations and propose identity-aware convolutional neural network by using an identity-sensitive contrastive loss amid the identity-related information learning process. %, ending up reaching a better accuracy rate. 
Zhang \emph{et al}. \cite{2018Joint} use generative adversarial network to alleviate pose-invariance along with simultaneous facial image synthesis and FER. Wang \emph{et al}. \cite{2019Identity} use adversarial feature learning method to reduce the disturbance caused by both facial identity and pose variations, yet those methods can only handle one or two disturbing factors. Ruan \emph{et al}. \cite{2020Ruan} propose disturbance-disentangled learning method to simultaneously disentangle multiple disturbing factors by taking advantage of multi-task learning and adversarial transfer learning.
\vspace{0.2cm}
\par
\textbf{Self-Paced Learning.} Self-paced learning (SPL) is a recently proposed methodology designed by mimicking the learning rules of humans. It processes training data from easy to hard step by step. As a consequence, during the learning process the model can be built from general to specific, thus can better fit the data and not be misguided easily \cite{Kumar2010Self}. Various degrees of success have been achieved using variants of SPL methods. Ma \emph{et al}. propose self-paced co-training in which SPL is applied to multi-view or multi-modality problems \cite{2017Fan}. In \cite{2017SelfMixture}, Han \emph{et al}. propose a novel self-paced regularizer for mixture of regressions.
%poorly conditioned linear sub-regressors was ameliorated by the effort of Han et al on the field of regression mixture with SPL. 
Zhao \emph{et al}. \cite{2015curriculum} realize a more effective real valued (soft) weighting manner from the original conventional binary (hard) weighting scheme for SPL. In \cite{2020ren} and \cite{2017ren}, Ren \emph{et al}. introduce soft weighting schemes of SPL to reduce the negative influence of outliers and noisy samples. In \cite{Huang0P021}, Huang \emph{et al.} propose a non-linear fusion SPL method for multi-view clustering to reduce the negative impact caused by variation of characteristics and qualities of different views. In \cite{ijms23073900}, Zhao \emph{et al.} introduce a single cell self-paced clustering for single cell RNA sequencing. In computer vision, deep learning methods usually achieve remarkable performance but also suffer from the non-convexity of the solution space and data noisy issues. Thus, SPL has also been widely used in deep neural networks. For instance, Li \emph{et al.} \cite{2017Li} seek to enhance the learning robustness of CNNs with SPL and propose SP-CNNs, in which the imbalance of training data in the discriminative model can be omitted. 
However, to our knowledge, the usage of SPL in FER tasks has not been well explored except \cite{ShaoWLHP022}.

\begin{figure*}[ht]
    \centering
    \includegraphics[scale=0.28]{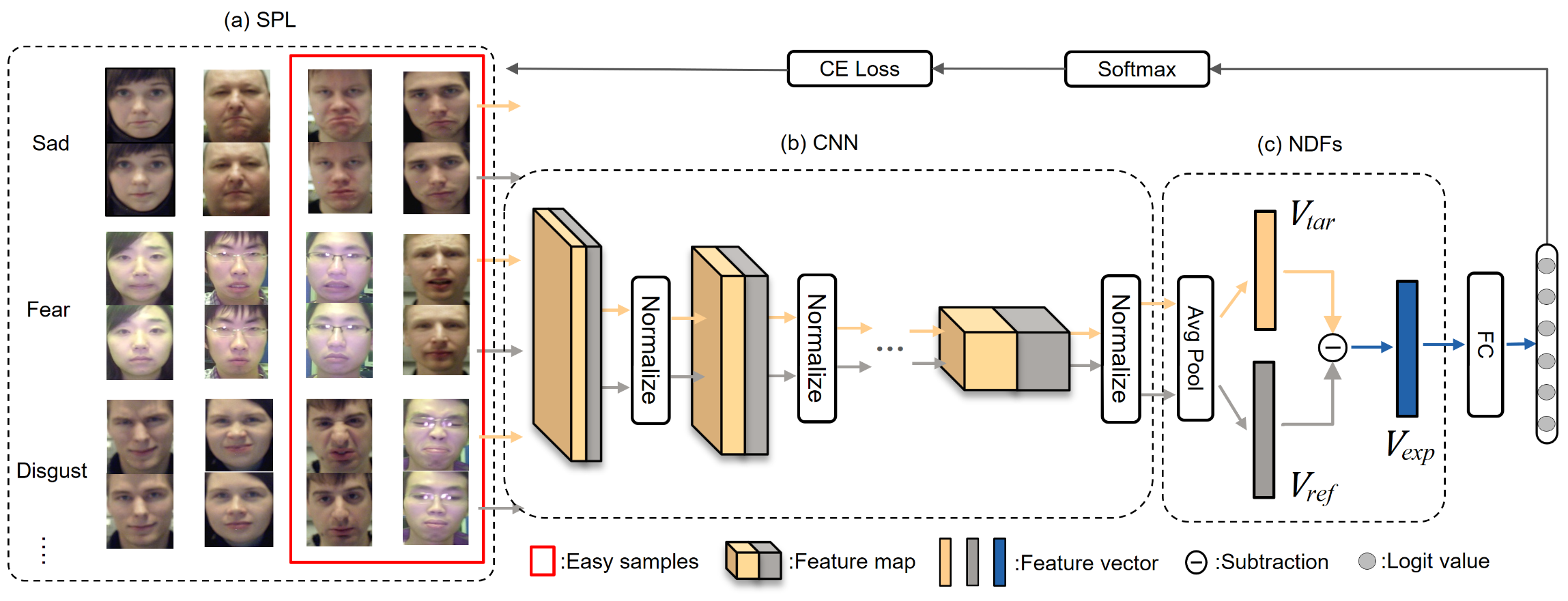}
    \caption{Overview of our proposed method. (a) Self-paced Learning (SPL). Target images with their reference images are presented in different categories, easy samples are chosen by categories to participate in training stage. (b) The backbone CNN. Target images and reference images are input to the backbone network simultaneously to obtain paired basic feature vectors. In each layer of the CNN, a normalization operation is applied on the paired feature maps. (c) Neutral expression-disentangled features (NDFs). NDFs are obtained by a subtraction operation. NDFs contain expression deviation information and few disturbance information. The cross entropy (CE) loss value of each paired sample is further used for SPL.}
    \label{fig:Network}
\end{figure*}

\section{Proposed Method}

\subsection{Overview}
Figure \ref{fig:Network} shows an overview of the proposed SPNDL.
SPL first selects a part of paired samples (target image and reference neutral image) that are easy for the initial network. Then, paired feature vectors of these chosen samples are extracted by a backbone CNN. In each layer, a normalization operation is applied on the paired feature maps to preliminarily eliminate the distribution difference. After this, a simple but very effective subtraction operation is applied on the paired feature vectors to obtain NDFs, in which disturbance information is removed and deviation information is collected. Finally, NDFs are used for classification and samples' loss values are used for the next pace of SPL.      

\subsection{Neutral Expression-Disentangled Features}
\label{sec:NDFs}
Information in facial expression image consists of expression-related information and expression-unrelated information (disturbance information). The disturbance information can confuse networks in training stage and make micro-facial movements more difficult to perceive. Let $\bm{V}_{tar}$ be the basic feature vector of a target image. We assume that there is an ideal method to decompose $\bm{V}_{tar}$ into expression feature vector $\bm{V}_{exp}$ and disturbance feature vector $\bm{V}_{dis}$:
\begin{equation}
    \bm{V}_{tar} = \bm{V}_{exp} + \bm{V}_{dis}.
\label{eq:vexp2}
\end{equation}
But in practice, such an ideal decomposition is hard to achieve. Fortunately, we find that neutral information in neutral facial expression consists of initial state of expression and similar disturbance information shared with its corresponding target image. With the ideal decomposition method, the basic feature vector of a reference neutral image $\bm{V}_{ref}$ can be decomposed into initial state feature vector $\bm{V}_{init}$ and disturbance feature vector $\bm{V}_{dis}$: 

\begin{equation}
    \bm{V}_{ref} = \bm{V}_{init} + \bm{V}_{dis}.
\label{eq:vexp1}
\end{equation}

Based on the above-mentioned observations, instead of directly decomposing $\bm{V}_{tar}$ and $\bm{V}_{ref}$, we use a subtraction operation to remove the interference of disturbance information and obtain the initial state of expression, which can help networks to extract deviation information. Thus the neutral expression-disentangled features $\bm{V}_{exp-d}$  focused on expression deviation information is given by
\begin{equation}
    \bm{V}_{exp-d} = \bm{V}_{tar}-\bm{V}_{ref}=\bm{V}_{exp}-\bm{V}_{init}.
\label{eq:vexp3}
\end{equation}

This formulation is based on early analysis of facial expression manifold, which shows the variations of face images can be represented as low dimensional manifolds in feature space and individual facial expression images can always gather around close to the neutral face in a sub-manifold \cite{2021Zhang, YaChang, 1978Facial, 2005Shan}. Therefore, it is reasonable to obtain NDFs by the proposed subtraction operation.     

%image sequences of facial expressions of an individual makes a continuous manifold. However, due to significant appearance variation across different subjects, the manifolds of different subjects vary much in the covered regions and the stretching directions. Those expressions from the same identity always Besides, the semantic-similar expressions from different identities are analogous on the expression manifold.
%DLN\cite{2021Zhang} also used such a disentangled manner, they fixed a pre-trained FaceNet's parameters to get identity feature vector and then stripped it from the output feature vector of a training FaceNet. Their method relies too much on the effect of pre-training, and fixing parameters leads to that the expression information is also included in the identity feature vector. Our method uses a shared network to extract $\bm{V}_{tar}$ from target image and $\bm{V}_{ref}$ from reference image at the same time. Target expression information only exists in $\bm{V}_{tar}$ while initial state information of expression only exists in $\bm{V}_{ref}$. And they share similar disturbance information.
\par
However, when there is a large distribution difference between the features of target image and reference image, the proposed subtraction operation will make the generated $\bm{V}_{exp-d}$ meaningless or even harmful. Thus we apply a normalization operation, which is performed based on the BN layer of ResNet-18 but plays a different role -- conducting normalization between the output features of target image and reference image in the backbone CNN's each layer. Let $\bm{F}_{tar}^{l}$ be the output features of target image in the $l^{th}$ convolutional layer and $\hat{\bm{F}}_{tar}^{l}$ be the input features. Let $\bm{F}_{ref}^{l}$ be the output features of reference image in the $l^{th}$ convolutional layer and $\hat{\bm{F}}_{ref}^{l}$ be the input features. We first calculate average value $\mu$ and variance $\sigma^{2}$  by\footnote{For simplicity, the $i^{th}$ $\bm{F}_{tar}^{l}$ (or $\bm{F}_{ref}^{l}$) is denoted by $\bm{F}_{tar}^{l}$ (or $\bm{F}_{ref}^{l}$).}
\begin{equation}
    \mu=\frac{1}{M} \sum_{i=1}^{M}(\bm{F}_{tar}^{l}+\bm{F}_{ref}^{l}),
\label{eq:normalization1}
\end{equation}

\begin{equation}
\sigma^{2} = \frac{1}{M} \sum_{i=1}^{M}\left((\bm{F}_{tar}^{l}-\mu)^{2} + (\bm{F}_{ref}^{l}-\mu)^{2}\right),
\label{eq:normalization2}
\end{equation}
where $M$ represents the number of paired samples in a batch. Then, we obtain the input features of the next layer by 

\begin{equation}
\hat{\bm{F_{.}}}^{l+1} = g\left(\gamma \frac{\bm{F_{.}}^{l}-\mu}{\sqrt{\sigma^{2}+\epsilon}} +\beta\right),
\label{eq:normalization3}
\end{equation}
where $\bm{F_{.}}^{l}$ represents $\bm{F}_{tar}^{l}$ or $\bm{F}_{ref}^{l}$, $\hat{\bm{F_{.}}}^{l}$ represents $\hat{\bm{F}}_{tar}^{l}$ or $\hat{\bm{F}}_{ref}^{l}$,  $\gamma$ and $\beta$ are reconstruction parameters learned by the networks, $\epsilon$ is a constant that prevents the denominator from being zero, and $g(\cdot)$ denotes activation function. Eq. (\ref{eq:normalization1}) to Eq. (\ref{eq:normalization3}) are all calculated separately for each channel of the feature maps. At last, the basic feature vector of target image or reference image (represented by $\bm{V_{.}}$) can be obtained by a average pooling implemented on the last layer's output features (represented by $\bm{F_{.}}^{-1}$):
\begin{equation}
\bm{V_{.}} = Average Pooling(\bm{F_{.}}^{-1})
\end{equation}

\par
\begin{figure}[!t]
\centering
{\includegraphics[scale=0.25]{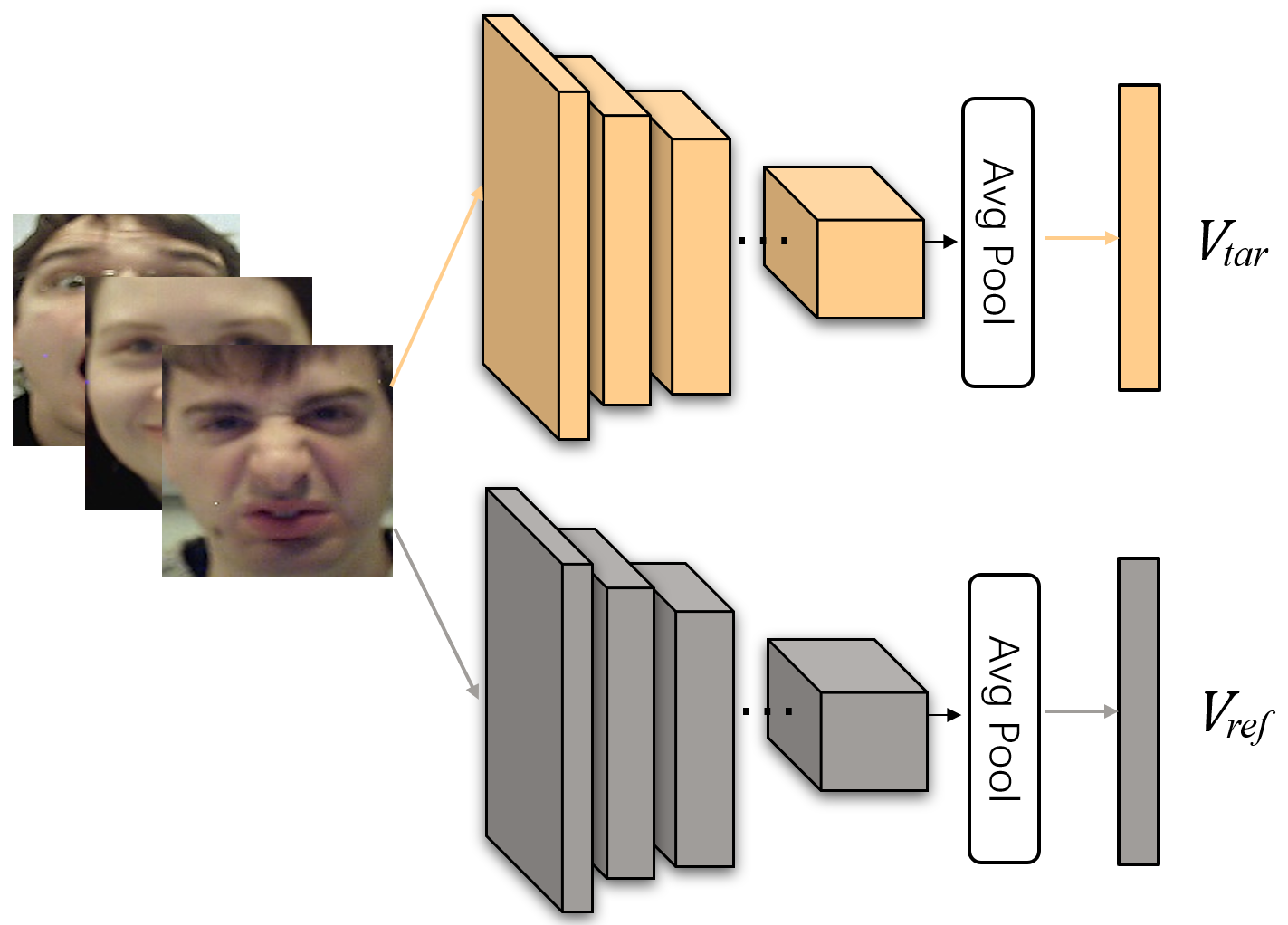}}
 \caption{Basic fearture extracting without neutral expression images. The two backbone convolutional neutral networks are both trainable. The yellow one is the expression-information-extracted network while the grey one is the neutral-information-extracted network.}
 \label{fig:new}
\end{figure}

One may argue that in real life it is possible that for a certain facial expression image, there is no corresponding neutral expression image to match it. We here propose a solution that can extract neutral information from the recognized images themselves. Learning from deviation learning network (DLN) \cite{2021Zhang}, we use two backbone CNN architectures, one extracts expression information and the other extracts neutral information. Figure \ref{fig:new} shows this basic feature extraction part, and the other parts are the same as our proposed method. Therefore, without a reference neutral image as input, the neutral expression-disentangled features can be obtained by
\begin{equation}
    \bm{V}_{exp-d} = f(x_{tar}) - f'(x_{tar}) = \bm{V}_{tar} - \bm{V}_{ref},
\end{equation}
where $f(\cdot)$ represents the expression-information-extracted network and $f'(\cdot)$ represents the neutral-information-extracted network. 

The neutral-information-extracted network can be trained while DLN uses a fixed CNN to obtain target images' identity information, which relies too much on pre-training. The expression-information-extracted network and the neutral-information-extracted network are both fine-tuned with the same ResNet-18 model, which is trained on the MS-Celeb-1M face recognition dataset. The effect of our subtraction operation is to weaken some parts of $V_{tar}$ (subtraction of numbers with the same sign) or enhance some parts of $V_{tar}$ (subtraction of numbers with different signs). At the beginning of training, $V_{tar}$ and $V_{ref}$ are almost equal. To make the subtraction effective, the model will learn which parts of $V_{tar}$ need to be weakened and which parts need to be enhanced. In other words, the model learn by itself to obtain expression-related information while removing expression-unrelated information.
\par

\subsection{Self-Paced Neutral Expression-Disentangled Learning}
Although normalization operation can alleviate the distribution difference to some extent, it is still difficult to generate consistently distributed NDFs from some paired samples. Here we propose a SPL strategy to further suppress the impact of this problem.
\par

With our SPL strategy, the training process of the network is divided into multiple paces. In the early stages of training, with large learning rate, the network is in an unstable exploratory state, which means the network can be easily impacted. Therefore, rather than considering all the samples simultaneously, we give priority to samples whose NDFs are safe and easy to the networks for training at early paces. 
As the network becomes robust, those samples excluded in early paces will be gradually included. At the last pace, the network will be trained with all of the samples.    
In addition, we found that in some databases, certain classes of samples are more difficult to learn than samples in other classes, resulting in an unfair selection of samples in SPL process. For example, in the experiments we found that most excluded samples are “fear” faces and “angry” faces, which in turn lead to the network learning unbalanced sample distribution in early paces. Therefore, instead of just simply selecting easy ones from all of the samples, we independently select samples within each class to ensure that the selection in each class tends to be consistently proportional.
\par
Let $\mathcal{X} = \{(\mathbf{x}_{tar}^{(1)}, \mathbf{x}_{ref}^{(1)}, y^{(1)}) , \cdots,(\mathbf{x}_{tar}^{(N)}, \mathbf{x}_{ref}^{(N)}, y^{(N)})\}$ be the data set where $\mathbf{x}_{tar}^{(i)} \in \mathbb{R}^{W \times H}$ is the $i^{th}$ target image, $\mathbf{x}_{ref}^{(i)} \in \mathbb{R}^{W \times H}$ is the $i^{th}$ reference image, $N$ is the total number of samples, and $y^{(i)} \in\{1, 2, \cdots, K\}$ represents its label where $K$ is the number of classes ($K \geq 2$).  

Particularly, a latent weight variable ${v}_{k,i}$ is defined to indicate whether the $i^{th}$ sample of the $k^{th}$ class is selected or not. Depending on the degree of training complexity, ${v}_{k,i}$ will be optimized as $1$ (selected) or $0$ (unselected). The aim of self-paced learning is to learn model parameters $\bm{\theta}$ and the latent weight variable $\bm{v}$ simultaneously by minimizing:

\begin{equation}
\min _{\boldsymbol{\theta}, \bm{v}} \sum_{k=1 }^{K} \sum_{i\in N_{k}}{v}_{k,i} l_{k,i}-\sum_{k=1 }^{K}\lambda_{k}\sum_{i\in N_{k}}{v}_{k,i},
\label{eq:self-paced}
\end{equation}
where $N_{k}$ denotes the set of indices of samples in the $k^{th}$ class, $\lambda_{k}$ is a parameter defined by category controlling the learning pace, and $l_{k,i}$ is the CE loss of the $i^{th}$ sample in the $k^{th}$ class:  
%\begin{equation}
%\begin{split}
%l_{k,i}=L(y^{(i)}=k,g(\mathbf{x}_{tar}^{(i)}, %\mathbf{x}_{ref}^{(i)};\boldsymbol{\theta}))= -\log %\frac{e^{\boldsymbol{\theta}_{k}^{T} \bm{V}_{exp}^{(i)}}}{\sum_{l=1}^{K} %e^{\boldsymbol{\theta}_{l}^{T} \bm{V}_{exp}^{(i)}}}
%\end{split}
%\end{equation}
\begin{equation}
l_{k,i}=L(y^{(i)}=k,g(\mathbf{x}_{tar}^{(i)},
\mathbf{x}_{ref}^{(i)};\boldsymbol{\theta}))= -\log(p_{k,i}).
\end{equation}
Here, $p_{k,i}$ is the probability that the $i^{th}$ sample belongs to the $k^{th}$ class, defined as:
\begin{equation}
p_{k,i} = \frac{e^{\boldsymbol{\theta}_{k}^{T} \bm{V}_{exp-d}^{(i)}}}{\sum_{l=1}^{K} e^{\boldsymbol{\theta}_{l}^{T} \bm{V}_{exp-d}^{(i)}}},
\end{equation}
where $\boldsymbol{\theta}_{k}$ denotes the parameter vector of the $k^{th}$ class in the linear fully-connected layer, and $\bm{V}^{(i)}_{exp-d}$ defined in Section \ref{sec:NDFs} is the NDFs of the paired sample $\{ \mathbf{x}_{tar}^{(i)}, \mathbf{x}_{ref}^{(i)}\}$. 

Eq. (\ref{eq:self-paced}) can be viewed as a traditional classification problem with fixed $\bm{v}$. When $\theta$ is fixed, optimal $\bm{v}^{*}$ can be easily obtained using the following equation:
%shown using calculation by
\begin{equation}
{v}_{k,i}^{*}= 
\begin{cases}
1, & \text { if } l_{k,i}<\lambda_{k} \\ 
0, & \text { otherwise }\end{cases}.
\label{eq:solve_v}
\end{equation}
Paired samples whose loss values are less than $\lambda_{k}$ will be declared as easy samples and the corresponding $v_{k,i}$ will be set to one. These paired samples are chose to train the network, while the others are excluded. $\lambda_{k}$ iteratively increases between paces, samples in each class are dynamically and proportionally involved in the training process, starting with easy examples and ending up with all samples. 
\par
Aiming at minimizing Eq. (\ref{eq:self-paced}),  the model parameters obtained at each pace will be used for initialization at the next pace (where the networks are retrained with the new chosen samples). 
In this way, those easy samples with high qualities (which are selected at early paces) contribute more to the training process, while the negative impact of those low-quality samples (such as ambiguous samples) and samples with inconsistently distributed NDFs can be reduced.
%our model is initialized to the result of the last pace when a new pace starts, then the networks are retrained with the new chosen paired samples. As such, our model always starts learning with the result of the previous learning pace, adaptively calibrated by easy samples. In this process, not only low-quality samples (such as ambiguous samples), but also samples with inconsistently distributed NDFs are excluded.

\subsection{Optimization}    
The alternative search strategy is adopted to solve Eq. (\ref{eq:self-paced}). We iteratively optimize the two parameters $\boldsymbol{\theta}$ and $\bm{v}$, \emph{i.e.}, updating one with the other being fixed. The main procedure consists of the following two steps.
\par
\noindent
\textbf{Optimizing $\bm{v}$ with fixed $\boldsymbol{\theta}$.} $v_{k,i}$ is a binary variable that indicates the $i^{th}$ sample of the $k^{th}$ class is selected or not. %When optimizing it, $\boldsymbol{\theta}$ is fixed. 
Given $\lambda_{k}$, we can directly use the Eq.~(\ref{eq:solve_v}) to obtain the optimal $v_{k,i}$.
At the first pace of the training stage, the parameter $\lambda_{k}$ can be initialized to exclude $50\%$ hard samples in each class. Then, at each subsequent pace, $\lambda_{k}$ is progressively increased to involve $10\%$ more samples until all the samples are selected. %In this learning process, the network gradually approaches to the optimal solution, alleviating the non-convex optimization problem.
\par
\noindent
\textbf{Optimizing $\boldsymbol{\theta}$ with fixed $\bm{v}$.} With samples chosen by the fixed $\bm{v}$, the network parameters $\boldsymbol{\theta}$ are updated through gradient descent. In early stages, these chosen samples are high-quality and easy to learn, so $\boldsymbol{\theta}$ converges easily.

%\subsection{Complexity Analysis}
%In the $l^{th}$ layer of CNN, respectively, $M_{l}$, $K_{l}$ and $C_{l}$ are denoted as the feature map size, the convolution kernel size, and the channel number. We also denote $N$ as the data size, $n$ as the batch size, $M$ as the number of categories, $D$ as the number of layers of CNN, and $H$ as the dimensionality of NDFs. In the mini-batch optimization process, the complexity of the FC layer is $O(nMH)$, the complexity of computing the cross-entropy loss is $O(nM)$. Moreover, the complexity of CNN is $O(n\sum_{l=1}^{D}M_{l}^{2}K_{l}^2C_{l-1}C_{l})$. Therefore, the total complexity to our train model is $O(N/n(nMH + nM + n\sum_{l=1}^{D}M_{l}^{2}K_{l}^2C_{l-1}C_{l})) = O(N \times \mbox{\emph{a constant value}})$. In conclusion, the complexity of SPNDL is linear to the data size $N$.    

\section{Experiments}
We adopt three variations of our proposed method for experiments: baseline (which only extracts features from target images using ResNet-18), SPNDL, and SPNDL w/o neutral image (which is described in Figure \ref{fig:new}).
\subsection{Experimental Setup}
\label{section:ExperimentalSetup}

\textbf{Databases.} The experiments are carried out on three popular databases: \textbf{Extended Cohn-Kanade Database (CK+)} \cite{2010ck+} contains 593 video sequences, recording 123 subjects' facial expressions in controlled lab environments. Each video sequence starts with the neutral expression and ends with the peak expression. There are only 327 video sequences with 118 subjects that are labeled as one of seven expressions, \emph{i.e.}, anger, contempt, disgust, fear, happiness, sadness, and surprise. We choose the first neutral frame and three peak expression frames from each video sequence. The first neutral frame is paired with each peak frame as a sample, thus resulting in a total of 981 paired samples. \textbf{Oulu-CASIA Database} \cite{2011oulu} captured 480 facial expressions from 80 subjects under three different illumination conditions using two types of cameras, thus containing a total of $480 \times 6$  video sequences. In our experiments, only the sequences captured under strong illumination condition with the VIS camera are used. Each of them is labeled as one of the six expressions (\emph{i.e.}, anger, disgust, fear, happiness, sadness, and surprise). Similar to the experimental setting on the CK+ database, from each video sequence, the first neutral frame is paired with each of the three peak expression frames to construct three paired samples, resulting in a total of 1440 paired samples. \textbf{Real-world Affective Faces Database (RAF-DB)}\cite{rafdb} is a in-the-wild database, containing 30,000 images labeled with basic or compound expressions by 40 trained human coders. 12271 training images and 3068 testing images with seven expressions (\emph{i.e.}, neutral, happiness, surprise, sadness, anger, disgust, and fear) are used in our experiments. Because there is no reference neutral image for each target image, RAF-DB is only used in experiments without neutral images.
\par 
\textbf{Comparison methods.}
We compare our method with both sequence-based methods and image-based methods on the CK+ and Oulu-CASIA databases. Specifically, sequence-based methods evaluate on the whole facial expression video sequences, the comparison methods include:   
\vspace{-0.2cm}
\begin{itemize}[leftmargin=*]
    \item Learning expressionlets on spatio-temporal manifold (STM-Explet) \cite{2014Liu}    
    \item Deep temporal appearance-geometry network (DTAGN) \cite{2015Joint}
    \item Dynamic FER using longitudinal facial expression atlases (Atlases) \cite{2012Atlases}
\end{itemize}
\vspace{-0.2cm}
\par
Rather than using the whole video sequences, the following image-based methods only evaluate on the three peak frames of each sequence:
\vspace{-0.2cm}
\begin{itemize}[leftmargin=*]
     \item Deep disturbance-disentangled learning (DDL) \cite{2020Ruan}
    \item De-expression residue learning (DeRL) \cite{2018Yang}
    \item Peak-piloted deep network (PPDN) \cite{2016Peak}
    \item Feature decomposition and reconstruction learning (FDRL) \cite{2021Feature}
    \item Inconsistent pseudo annotations to latent truth (IPA2LT) \cite{2016FaceNet2ExpNet}
    \item Region attention network (RAN) \cite{RAN}
    \item Selfcure network (SCN) \cite{2020Suppressing}
    \item Facial expression recognition with grid-wise attention and visual transformer (FER-VT) \cite{FER-VT}
    \item Harmonious Representation Learning (HRL) \cite{HAN2022104}
\end{itemize}
\vspace{-0.2cm}
\par
\textbf{Evaluation metrics.} The performance of all the compared methods is evaluated by the recognition accuracy (ACC). Because both CK+ and Oulu-CASIA databases are not divided into training set and test set, following the compared methods, a 10-fold subject-independent cross-validation is performed. The averaged  results of 10-fold cross-validation are reported.
%The ACC is the average of the 10-fold experiments' results.

\subsection{Implementation Details}
On CK+ and Oulu-CASIA databases, the face region of each facial image is cropped by four boundary points of the landmark points (left boundary point, right boundary point, upper boundary point and lower boundary point). On RAF-DB, the facial images are all detected and aligned using Retinaface \cite{detected2020}. All images are resized to $224\times224$ pixels on each database. During the training process, a random horizontal flipping or Gaussian noise adding operation is applied for data augmentation purposes to increase the number of images to avoid overfitting. Notice that the target image and reference image in each paired sample are augmented by the same operation simultaneously. 
\par 
The SPNDL method is implemented with the Pytorch toolbox and the backbone is a ResNet-18 model \cite{2016resnet} pre-trained on the ImageNet database \cite{2009ImageNet}. On CK+ and Oulu-CASIA databases, we train SPNDL and SPNDL (w/o neutral image) in an end-to-end manner for 250 epochs in each run of the 10-fold experiments. On RAF-DB, because there is on reference neutral image, we only train SPNDL (w/o neutral image) for 150 epochs. The batch size is set to 128 for all databases. In the training stage, we adopt Adam as the optimization method, with learning rate $lr = 0.001$, first-order momentum $\beta_1 = 0.9$, and second-order momentum $\beta_2 = 0.999$. All experiments are run with Python 3.8 and Pytorch 1.7.1 on a Linux server equipped with 2.5GHz CPU and 16GB RAM, a single RTX3070 GPU is used to accelerate the training stage.

\begin{table}[t]
  \caption{Recognition accuracy ($\%$) on the CK+ for 7 expressions classification.}
  \centering
    \begin{tabular}{p{5.5cm}p{4.3cm}p{2.6cm}}
    \hline
    \multicolumn{1}{c}{Method} &
    \multicolumn{1}{c}{Setting} & 
    \multicolumn{1}{c}{Accuracy} \\
    \hline
    \makecell[r]{STM-E\cite{2014Liu}(2014)\hspace{0.9cm} }&  \makecell[c]{sequence-based}&
    \makecell[c]{94.19}\\
    \makecell[r]{DTAGN\cite{2015Joint}(2015)\hspace{0.9cm} } & \makecell[c]{sequence-based} & \makecell[c]{97.25}  \\
    \makecell[r]{PPDN\cite{2016Peak}(2016)\hspace{0.9cm} }&  \makecell[c]{image-based}&
    \makecell[c]{99.30}\\
    \makecell[r]{DeRL\cite{2018Yang}(2018)\hspace{0.9cm} } & \makecell[c]{image-based} & \makecell[c]{97.30}  \\
    \makecell[r]{DDL\cite{2020Ruan}(2020)\hspace{0.9cm} } & \makecell[c]{image-based} & \makecell[c]{99.16}  \\
    \makecell[r]{FDRL\cite{2021Feature}(2021)\hspace{0.9cm} } & 
    \makecell[c]{image-based} & \makecell[c]{99.54}\\
    \makecell[r]{HRL\cite{HAN2022104}(2022)\hspace{0.9cm} } & 
    \makecell[c]{image-based} & \makecell[c]{98.90}\\
     \hline
    \makecell[c]{Baseline(ResNet-18)} & \makecell[c]{image-based} & \makecell[c]{97.55} \\ 
    \makecell[c]{SPNDL(w/o neutral image)} & \makecell[c]{image-based} & \makecell[c]{99.18} \\
    \makecell[c]{\textbf{SPNDL}} & \makecell[c]{image-based} & \makecell[c]{\textbf{99.69}}  \\
    \hline
    \end{tabular}%
  \label{tab:comparision_ck+}%
\end{table}%

\begin{table}[tbp]
\caption{Recognition accuracy ($\%$) on the Oulu-CASIA for 6 expressions classification.}
  \centering
    \begin{tabular}{p{5.5cm}p{4.3cm}p{2.6cm}}
    \hline
    \multicolumn{1}{c}{Method} &
    \multicolumn{1}{c}{Setting} & 
    \multicolumn{1}{c}{Accuracy} \\
    \hline
    \makecell[r]{Atlases\cite{2012Atlases}(2012)\hspace{0.9cm} }&  \makecell[c]{sequence-based}&
    \makecell[c]{75.52}\\
    \makecell[r]{DTAGN\cite{2015Joint}(2015)\hspace{0.9cm} } & \makecell[c]{sequence-based} & \makecell[c]{81.46}  \\
    \makecell[r]{FN2EN\cite{2016FaceNet2ExpNet}(2016)\hspace{0.9cm} }&  \makecell[c]{image-based}&
    \makecell[c]{87.71}\\
    \makecell[r]{DeRL\cite{2018Yang}(2018)\hspace{0.9cm} } & \makecell[c]{image-based} & \makecell[c]{88.00}  \\
    \makecell[r]{DDL\cite{2020Ruan}(2020)\hspace{0.9cm} } & \makecell[c]{image-based} & \makecell[c]{88.26}  \\
    \makecell[r]{FDRL\cite{2021Feature}(2021)\hspace{0.9cm} } & \makecell[c]{image-based} & \makecell[c]{88.26}\\
    \hline
    \makecell[c]{Baseline(ResNet-18)} & \makecell[c]{image-based} & \makecell[c]{87.57}  \\
    \makecell[c]{SPNDL(w/o neutral image)} & \makecell[c]{image-based} & \makecell[c]{88.54}  \\
    \makecell[c]{\textbf{SPNDL}} & \makecell[c]{image-based} & \makecell[c]{\textbf{90.14}}  \\
    \hline
    \end{tabular}%
  \label{tab:comparision_oulu}%
\end{table}%

\begin{table}[tbp]
\caption{Recognition accuracy ($\%$) on the RAF database for 7 expressions classification.}
  \centering
    \begin{tabular}{p{5.5cm}p{4.3cm}p{2.6cm}}
    \hline
    \multicolumn{1}{c}{Method} &
    \multicolumn{1}{c}{Setting} & 
    \multicolumn{1}{c}{Accuracy} \\
    \hline
    \makecell[r]{IPA2LT\cite{IPA2LT}(2018)\hspace{0.9cm} }&  \makecell[c]{image-based}&
    \makecell[c]{86.77}\\
    \makecell[r]{RAN\cite{RAN}(2020)\hspace{0.9cm} } & \makecell[c]{image-based} & \makecell[c]{86.90}  \\
    \makecell[r]{DDL\cite{2020Ruan}(2020)\hspace{0.9cm} } & \makecell[c]{image-based} & \makecell[c]{87.71}  \\
    \makecell[r]{SCN\cite{2020Suppressing}(2020)\hspace{0.9cm} } & \makecell[c]{image-based} & \makecell[c]{88.14}  \\
    \makecell[r]{FER-VT\cite{FER-VT}(2021)\hspace{0.9cm} }&  \makecell[c]{image-based}&
    \makecell[c]{88.26}\\
    \makecell[r]{\textbf{FDRL}\cite{2021Feature}(2021)\hspace{0.9cm} } & \makecell[c]{image-based} & \makecell[c]{\textbf{89.47}}\\
    \makecell[r]{HRL\cite{HAN2022104}(2022)\hspace{0.9cm} } & 
    \makecell[c]{image-based} & \makecell[c]{87.77}\\
    \hline
    \makecell[c]{Baseline(ResNet-18)} & \makecell[c]{image-based} & \makecell[c]{87.13}  \\
    \makecell[c]{SPNDL(w/o neutral image)} & \makecell[c]{image-based} & \makecell[c]{89.08}  \\
    \hline
    \end{tabular}%
  \label{tab:comparision_raf}%
\end{table}%

\begin{figure}[!t]
  \centering
  \begin{subfigure}{0.32\textwidth}
  \centering
    \includegraphics[scale=0.3]{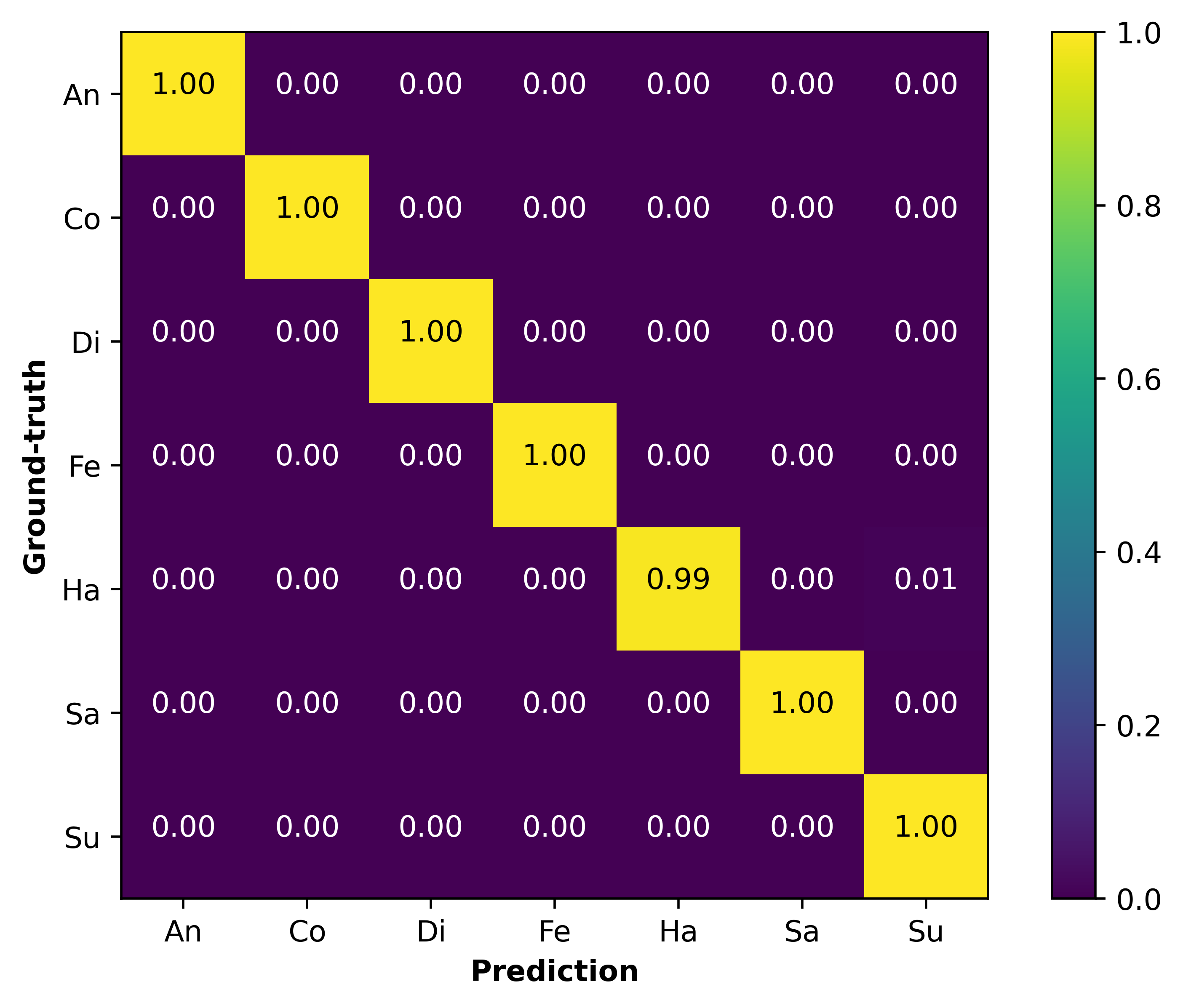}
    \caption{CK+}
    \label{fig:matrix_ck+}
  \end{subfigure}
  \hfill
  \begin{subfigure}{0.32\textwidth}
  \centering
    \includegraphics[scale=0.3]{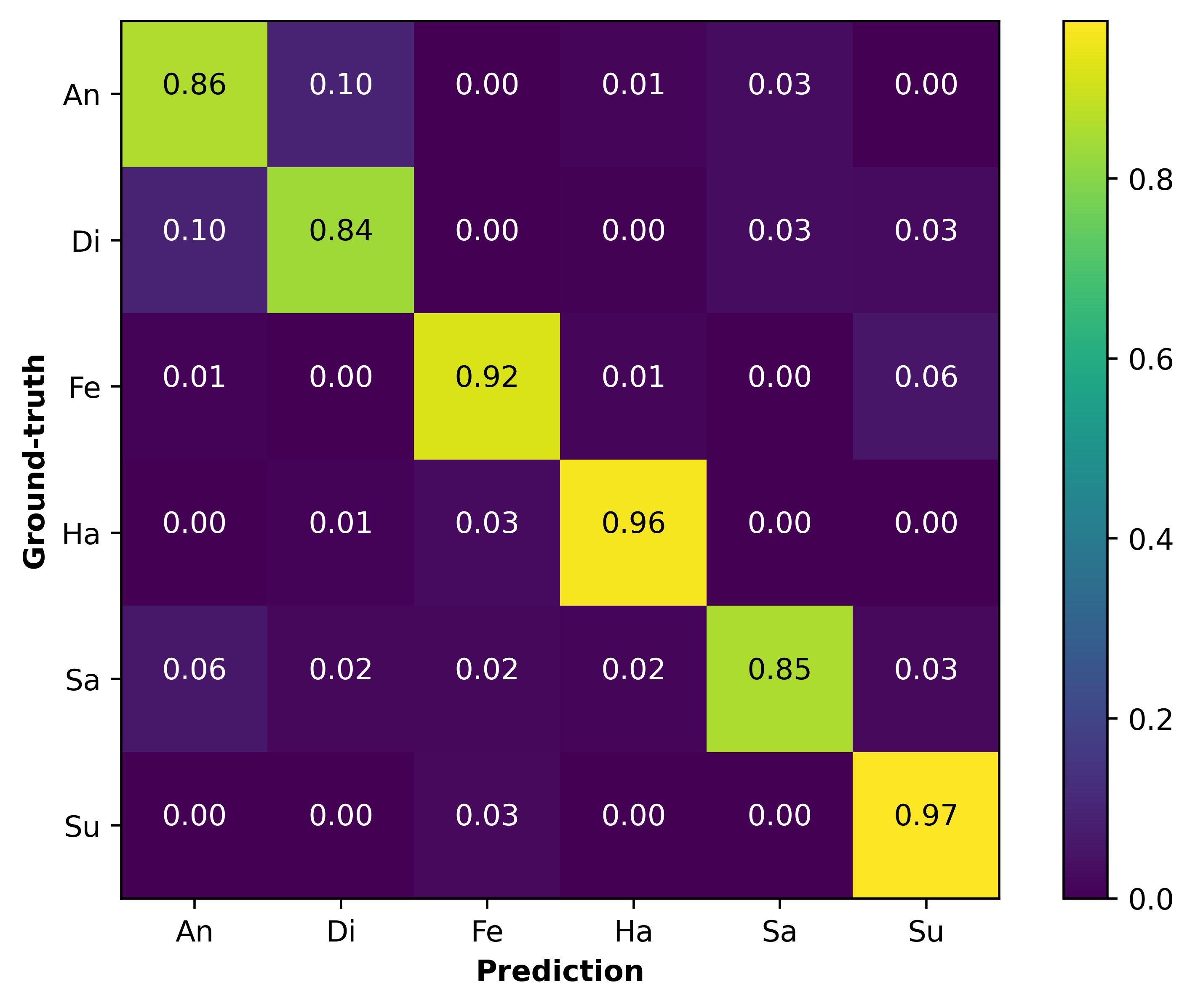}
    \caption{Oulu-CASIA}
    \label{fig:matrix_oulu}
  \end{subfigure}
  \hfill
  \begin{subfigure}{0.32\textwidth}
  \centering
    \includegraphics[scale=0.3]{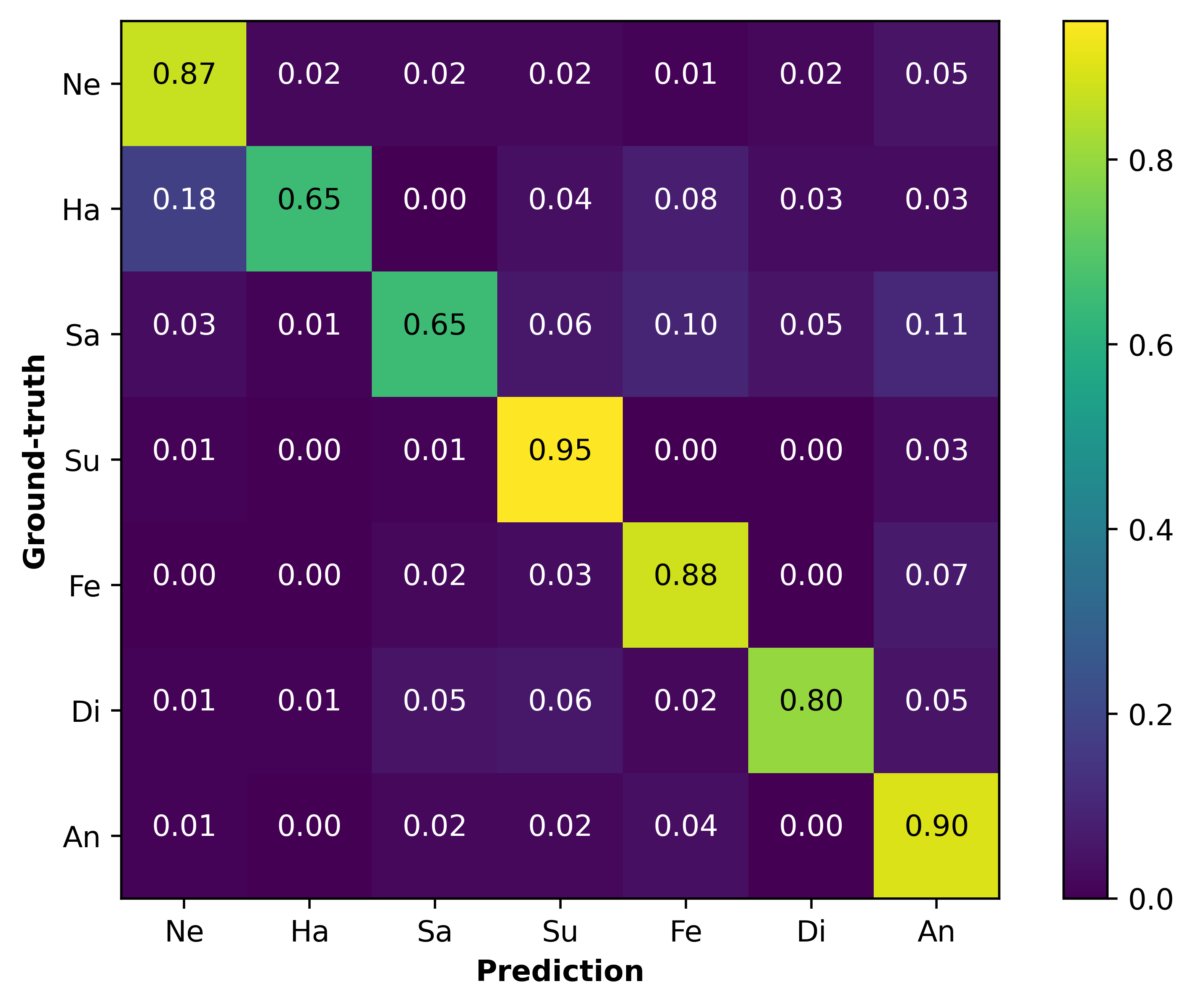}
    \caption{RAF-DB}
    \label{fig:matrix_raf}
  \end{subfigure}
  \caption{Confusion matrix of our SPNDL method on CK+, Oulu-CASIA and RAF-DB.}
  \label{fig:matrix}
\end{figure}
\par

\subsection{Comparison with State-of-the-Art}
We compare the recognition accuracy of the proposed SPNDL with the state-of-the-art methods mentioned in Section \ref{section:ExperimentalSetup} on CK+, Oulu-CASIA and RAF-DB. The corresponding results are reported in Table \ref{tab:comparision_ck+} , Table \ref{tab:comparision_oulu}, and Table \ref{tab:comparision_raf} respectively. The corresponding confusion matrix of our method is shown in Figure \ref{fig:matrix}. 
\par
On the two in-the-lab databases, our SPNDL outperforms the baseline method. Like other image-based methods, the performance of SPNDL is much better than those sequence-based methods. In particular, SPNDL gains a slight improvement of $0.15\%$ on CK+ compared with the recent state-of-the-art method FDRL. SPNDL outperforms FDRL and DDL by $1.88\%$ on Oulu-CASIA. It may be explained by the following reason. Many of the samples in the CK+ database are high-quality, with distinctive features that distinguish them from other facial expressions. Therefore, most methods can achieve high recognition accuracies of over $99\%$. However, there are many ambiguous samples and samples with micro-facial expressions in the Oulu-CASIA database. Our proposed method will exclude these difficult samples in training's early stages, helping the network avoid poor local solutions. With the gradual enhancement of extracting effective NDFs, the network is able to correctly classify these samples. Thus, our method can achieve a significant improvement on Oulu-CASIA.

\par
When there is no reference neutral image, experiments are conducted on RAF-DB, where most subjects have no neutral face. In particular, SPNDL (w/o neutral image) achieves recognition accuracy of $89.08\%$, outperforming most of the comparison methods. The accuracy is slightly lower than FDRL while SPNDL outperforms FDRL on CK+ and Oulu-CASIA. FDRL first decomposes expression basic features into a set of facial action-aware latent features, and then reconstruct the expression feature according to their intra-feature relation weights and their inter-feature relation weights, which are all learned by a network. Their method can find shared information among similar expressions while strengthen expression-specific variations. To find expression-specific variations, our method tries to obtain deviation information through the initial state of expression existing in neutral expression image. But initial state of expression is hard to obtain in in-the-wild datasets like RAF-DB. That’s the reason why our method only outperforms FDRL on in-the-lab datasets. However, our method still performs well on RAF-DB, because we can strengthen expression-related features by removing disturbance information.

To further prove the effectiveness of SPNDL (w/o neutral image), instead of inputting paired images, we remove the reference neutral images in CK+ and Oulu-CASIA. Our method achieves recognition accuracy of $99.18\%$ on CK+, which is much higher than the baseline method. On Oulu-CASIA, without neutral images, our method can still achieve state-of-the-art result. On both CK+ and Oulu-CASIA, it can be seen that the recognition accuracy without reference neutral images is better than that of baseline but worse than that with neutral images. This is because a part of the disturbance information in target image can be extracted to generate $\bm{V}_{ref}$ by the neutral-information-extracted network. However, at the same time, some useful expression information will inevitably be included in $\bm{V}_{ref}$, which will be removed from $\bm{V}_{tar}$ in the subtraction operation. Another reason is that expression's initial state cannot be obviously introduced, which means deviation information is hard to obtain.

\begin{table}
\caption{Ablation studies for two key methods of our model
on the CK+ and Oulu-CASIA databases. The recognition accuracy ($\%$) is used for performance evaluation.}
  \centering
    \begin{tabular}{p{1.3cm}p{1.7cm}|p{1.5cm}p{1.8cm}}
    \hline
    \multicolumn{1}{c}{\multirow{2}[2]{*}{NDFs}} & \multicolumn{1}{c|}{\multirow{2}[2]{*}{SPL}} & \multicolumn{1}{c}{\multirow{2}[2]{*}{CK+}} & \multicolumn{1}{c}{\multirow{2}[2]{*}{Oulu-CASIA}} \\
          &       &       &  \\
    \hline
    \makecell[c]{\XSolidBrush}  & \makecell[c]{\XSolidBrush} &  \makecell[c]{97.55}     & \makecell[c]{87.57}  \\
     \makecell[c]{\XSolidBrush} & \makecell[c]{\Checkmark} & 
 
    \makecell[c]{99.08}    &  \makecell[c]{88.33}\\
    \makecell[c]{\Checkmark} & \makecell[c]{\XSolidBrush} &  
    \makecell[c]{99.29}    &  \makecell[c]{89.58}\\
    \makecell[c]{\Checkmark} & \makecell[c]{\Checkmark} &   
    \makecell[c]{99.69}    &  \makecell[c]{90.14}\\
    \hline
    \end{tabular}%
 \label{tab:ablation}%
\end{table}%

\subsection{Ablation Studies}
The validity of our proposed method is mainly demonstrated by the following experiments on the CK+ and Oulu-CASIA databases. To get a reference standard result, we first feed all samples without neutral face into the backbone convolutional neutral network and directly use the output basic feature vector for classification task.
\par
\textbf{Neutral expression-disentangled features.} To validate the effectiveness of our NDFs, we feed all target images with their reference neutral images into the backbone to get paired basic feature vectors. Notice that target image and reference image are not fed into the same backbone CNN respectively, instead they are input simultaneously as members from one batch. So the normalization operation between them can be applied in each layer of the CNN. At last, a subtraction operation is implemented to obtain the final NDFs for classification.
\par
\textbf{Self-paced learning strategy.} Our SPL strategy consists of 6 paces. At the first pace, $\bm{\lambda}$ is initialized to obtain $50\%$ samples from each category to train the model. At the following paces, $\bm{\lambda}$ is progressively increased such that $10\%$ more samples were gradually involved at each pace. SPL stops when all the samples are included.

The experimental results of ablation studies are reported in Table \ref{tab:ablation}. It can be seen that separately incorporating NDFs or SPL into the backbone network can improve the performance, which shows the effectiveness of NDFs and SPL. Moreover, by employing NDFs and SPL simultaneously, we are able to achieve better recognition accuracy. This is because when SPL works alone, hard and low-quality samples are excluded. The network learns easy samples first and then gradually steps to hard ones, bad local solutions can be avoided in this process.

Specifically, when just using NDFs, the key features of facial expression are more easily to extract due to the elimination of disturbing factors. Deviation information is also included, enabling our model to capture distinguishing characteristics among similar expressions. When SPL and NDFs work together, they can be organically combined to reduce the impact of inconsistently distributed NDFs. Samples with these NDFs are excluded at early paces. At later paces, the network can gradually understand a part of these NDFs and is more likely to classify them correctly. When all the samples are included, meaningless NDFs will have limited negative impact because the network has become more robust. %cannot be optimized will have little impact because the network has become more robust.

\begin{figure}[!t]
\centering
{\includegraphics[scale=0.18]{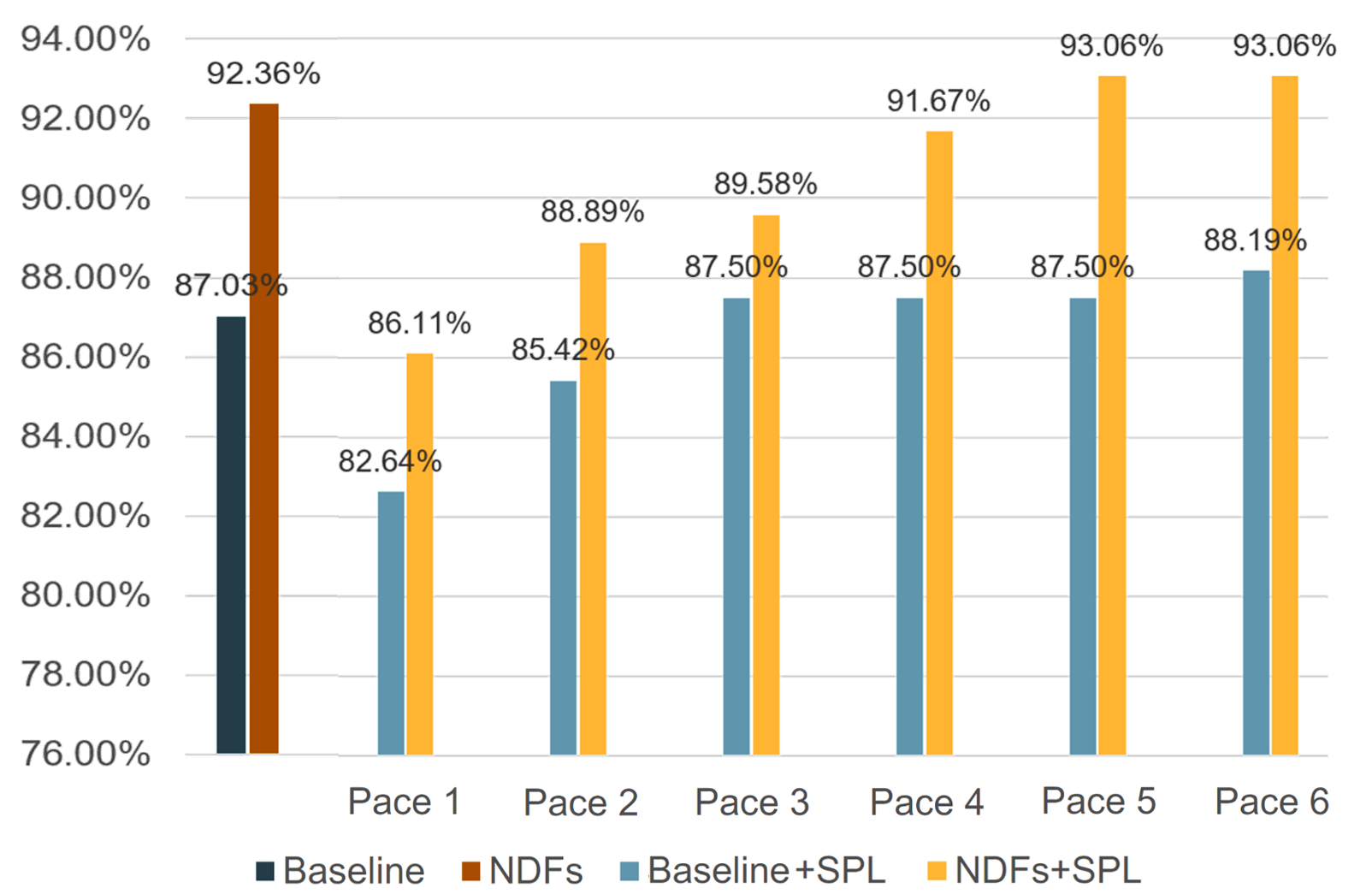}}
 \caption{The recognition accuracy at each pace of SPL with and without NDFs. The results are compared with the baseline and method only using NDFs.}
 \label{fig:compare}
\end{figure}

\par
Figure \ref{fig:compare} shows the recognition accuracy at each pace of SPL with and without NDFs. The results are from a random chosen experiment from the 10-fold experiments on the Oulu-CASIA database. In the first two paces, the accuracy of SPL is less than baseline (which is mainly because the training samples are not enough), but it achieves a better result at the third pace. The similar situation happens between method just using NDFs and method using NDFs $+$ SPL. The accuracy of the latter exceeds the former at the fifth pace. In addition, the performance of method with NDFs are always much better than method without NDFs, which proves the effectiveness of the proposed NDFs.

\begin{figure}[!t]
\centering
{\includegraphics[scale=0.6]{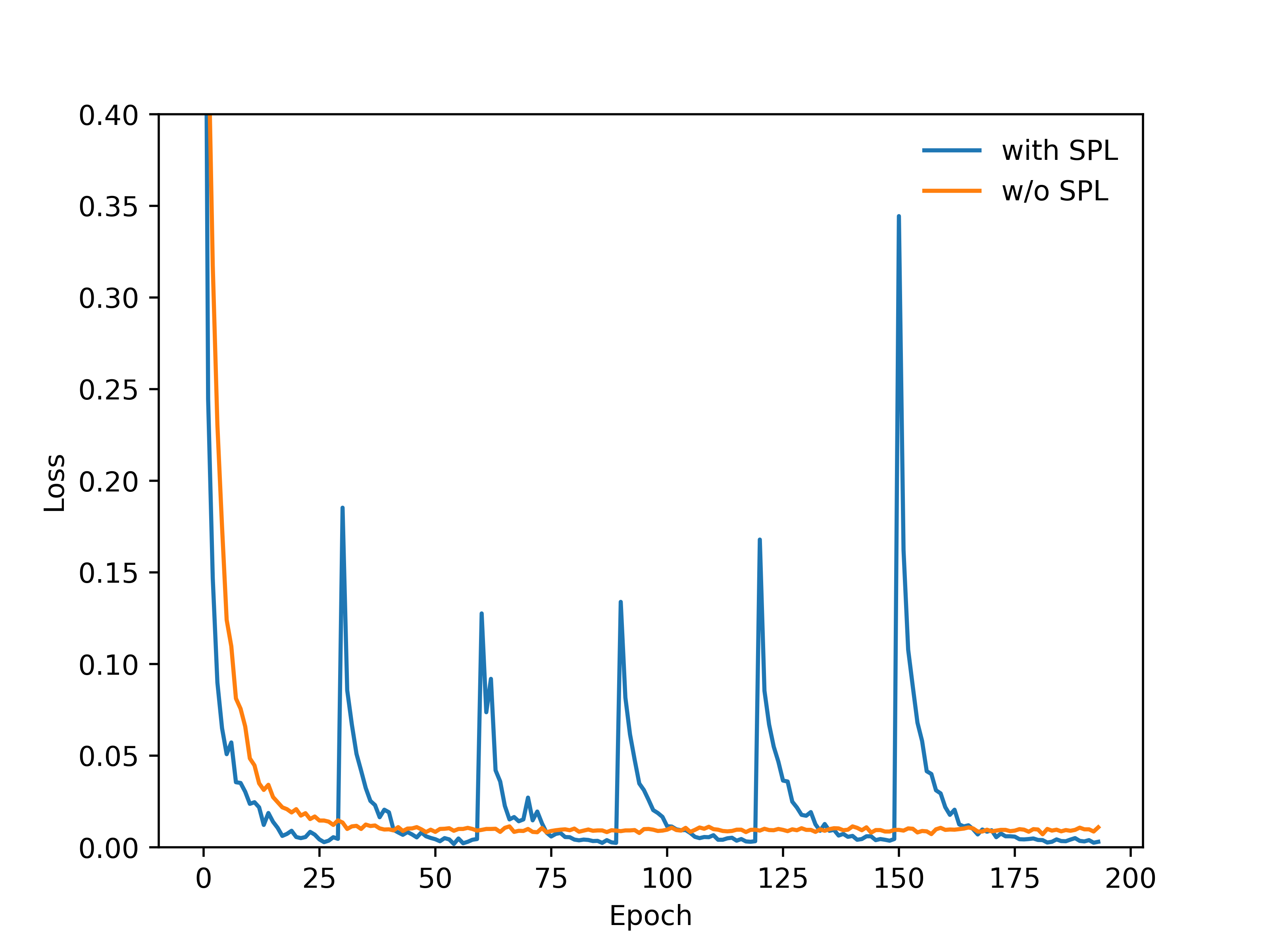}}
 \caption{Losses versus different epochs when training on RAF-DB. The blue line are training losses with SPL, and the yellow line are training losses w/o SPL.}
 \label{fig:loss}
\end{figure}

To prove the effectiveness of our SPL, we present the losses versus different epochs when training on RAF-DB with and w/o SPL in Figure \ref{fig:loss}. When training w/o SPL, the loss decreases gradually until it converges to about 0.005. When using SPL, there is a sudden increase of the loss in the beginning of each pace, this is because more training samples are added for training, but the loss still decreases gradually within a pace. At epoch 150, all samples are added for training, and the loss finally converges to about 0.003, which is lower than that of the method w/o SPL, showing our SPL can help the model escape from pool local solutions to a lower loss value.

\subsection{Visualization}
\begin{figure}[!t]
  \centering
  \begin{subfigure}{0.48\textwidth}
  \centering
    \includegraphics[scale=0.11]{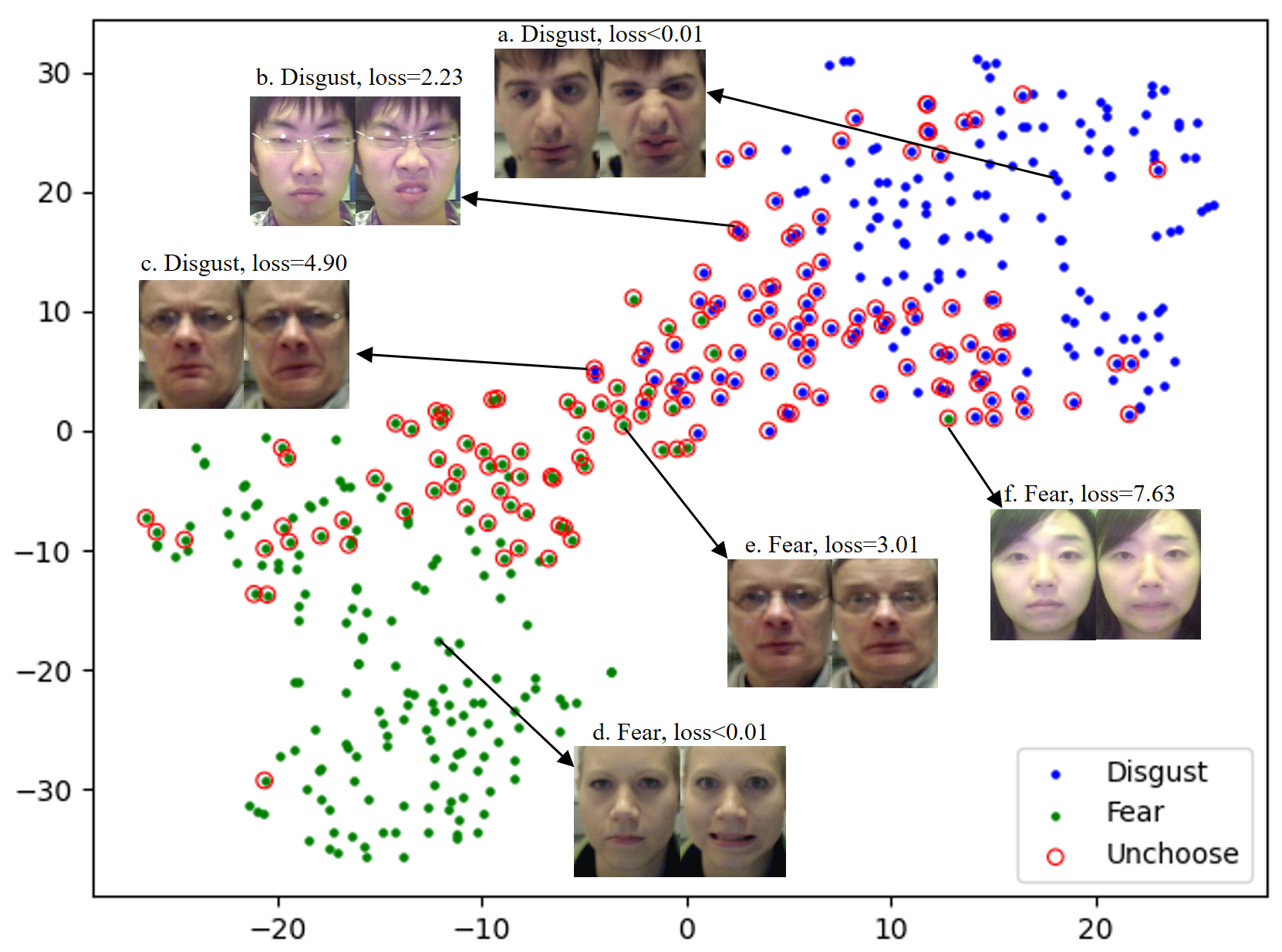}
    \caption{The distribution of NDFs at pace 2.}
    \label{fig:pace2}
  \end{subfigure}
  \hfill
  \begin{subfigure}{0.48\textwidth}
  \centering
    \includegraphics[scale=0.11]{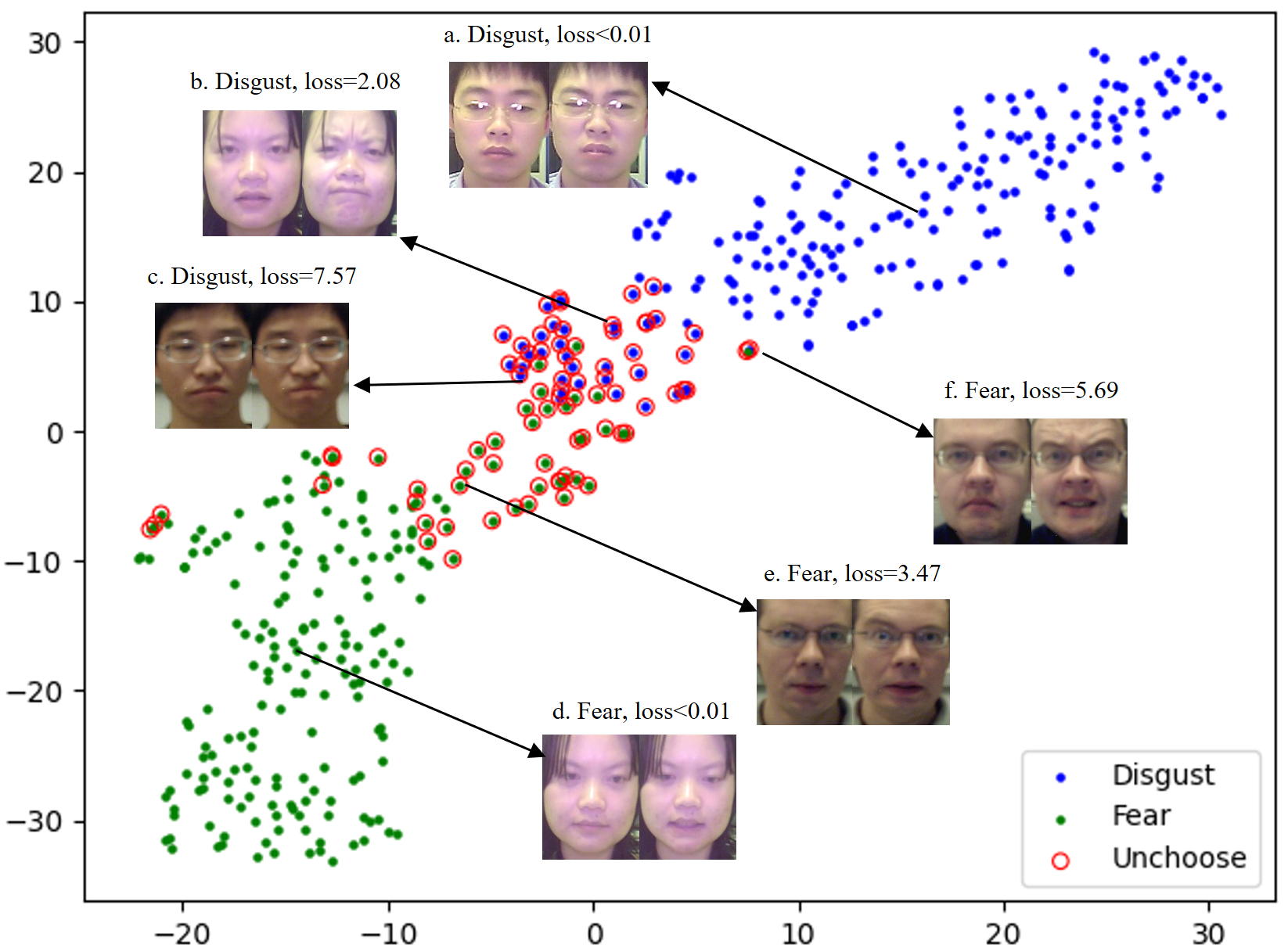}
    \caption{The distribution of NDFs at pace 5.}
    \label{fig:pace5}
  \end{subfigure}
  \caption{The distribution of NDFs from “fear” and “disgust” expressions in self-paced learning process. Pace 2 and pace 5 are randomly chosen for visualization. Three kinds of paired samples are chosen from each category to present. Circled paired samples are excluded for training.}
  \label{fig:distribution}
\end{figure}

\begin{figure*}[ht]
    \centering
    \includegraphics[width=13cm, height=5cm]{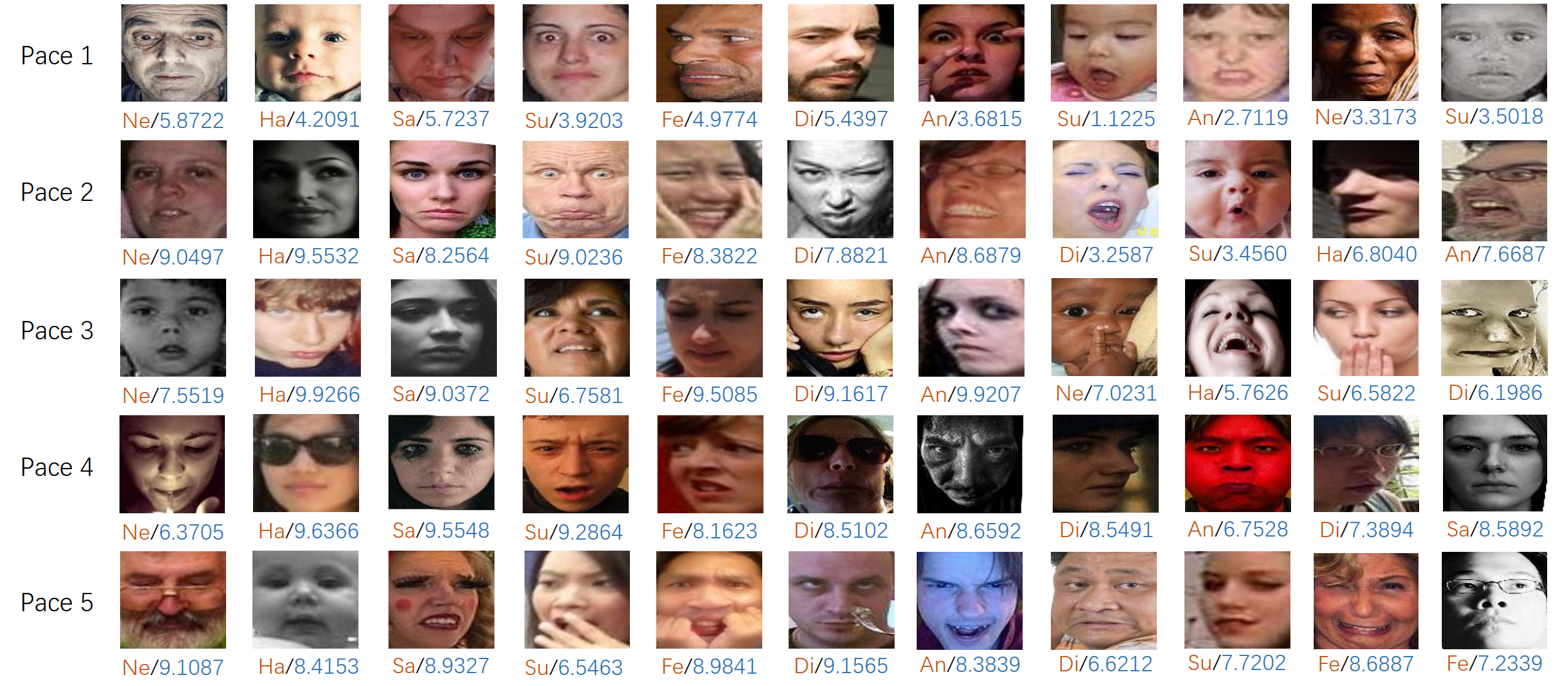}
    \caption{Samples excluded at each pace. We randomly choose some samples excluded by our method at each pace on RAF-DB. Corresponding labels (Ne=Neutral, Ha=Happiness, Sa=Sadness, Su=Surprise, Fe=Fear, Di=Disgust, An=Anger) and loss values are presented below the images. There is no pace 6 because all samples are included for training at pace 6.}
    \label{fig:unselected}
\end{figure*}

In Figure \ref{fig:distribution}, we visualize the distribution of NDFs via t-SNE \cite{van2008visualizing} from “fear” and “disgust” expressions at pace 2 and pace 5 on Oulu-CASIA. $40\%$ samples from each category are excluded at pace 2 while $20\%$ samples from each category are excluded at pace 5. There are three kinds of paired samples: samples located in the distribution center (samples a and d in Figure \ref{fig:pace2}, samples a and d in Figure \ref{fig:pace5}); samples farthest from the center (samples c and f in Figure \ref{fig:pace2}, samples c and f in Figure \ref{fig:pace5}); randomly chosen samples that have been excluded (samples b and e in Figure \ref{fig:pace2}, samples b and e in Figure \ref{fig:pace5}). The center samples have low loss values, they are easy to learn. Loss values of the last two kinds of samples are high, which is caused by micro-facial movements or similarities across different expressions. These samples are neglected at early paces. At pace 2, the inconsistently distributed problem of NDFs is obvious. At pace 5, the problem has been significantly improved.

Figure \ref{fig:unselected} shows some randomly chosen samples that were excluded when training on RAF-DB. Because these images are collected from the Internet, some of them are low-quality (\emph{e.g.}, occlusion, uneven illumination, and low resolution) while some are annotated with ambiguous or even incorrect labels. Training with these samples at early paces may resulting in poor local solutions. At pace 1, the loss values of poor samples are generally low, this is because the network cannot distinguish them well at the beginning of the training. However, these poor samples maintain high loss values at the following paces, which means they are prevented from affecting the network by SPL. At the last pace, all samples will be included for training to avoid overfitting.

\section{Conclusion and Future Work}
In this paper, we have proposed a self-paced neutral expression-disentangled learning (SPNDL) model, which is hyperparameter-free in the sense. Our method achieves state-of-the-art FER performance on the CK+ and Oulu-CASIA databases and competitive results on RAF-DB.
First, basic features of target image and the corresponding neutral image are extracted, both of which share similar disturbance information. To alleviate the inconsistently distributing problem of these basic features, we apply a normalization operation in each layer of the backbone CNN. Second, NDFs are obtained by a subtraction operation, which can suppress the negative impact of disturbing factors in facial expression images, and simultaneously introduce the information of expression's initial state. Finally, a SPL strategy is proposed in training stage, letting the network learn like human beings -- starting with easy NDFs and building up to complex ones. 

Our future work will focus on how to more effectively learn neutral expression-disentangled features when there are no corresponding neutral expression images, especially on large-scale in-the-wild databases where there are more disturbing factors.
%it is still hard to achieve state-of-the-art results.
%which means SPNDL does not work well on in-the-wild databases. 
%Thus, our future work will focus on extending our framework to large-scale databases where there are more disturbing factors and no corresponding neutral expression images.



\bibliographystyle{elsarticle-num} 
\bibliography{egbib}

\end{document}